\documentclass[letterpaper, 10 pt, conference]{ieeeconf}  

\usepackage{cite}
\usepackage{amsmath,amssymb,amsfonts}
\usepackage{algorithmic}
\usepackage{algorithm}
\usepackage{algorithmic}
\usepackage{graphicx}
\usepackage{textcomp}
\usepackage{xcolor} 
\usepackage{amsmath}
\usepackage{multirow}
\usepackage{svg}
\usepackage{subcaption}
\usepackage{graphicx}
\usepackage{makecell}
\usepackage{multirow}
\usepackage{booktabs}
\usepackage{comment}
\usepackage{graphicx}

\usepackage{balance}

\usepackage{caption}

\IEEEoverridecommandlockouts                              

\title{\LARGE \bf
A Real-Time System for Scheduling and \\ Managing UAV Delivery in Urban Areas}

\author{Han Liu, Tian Liu and Kai Huang
	\thanks{
		This work is supported by Ministry of Education Industry-University
		Collaborative Education Project under Grant 2408234204.
		H. Liu, T. Liu and K. Huang are with the School of Computer Science, Sun Yat-sen University, Guangzhou, China. \textit{(Corresponding author: Kai Huang)}}%
	\thanks{E-mail: liuh386@mail2.sysu.edu.cn,  huangk36@mail.sysu.edu.cn}
}

\begin{document}

\captionsetup[figure]{font=small}

\maketitle
\thispagestyle{empty}
\pagestyle{empty}

\begin{abstract}

As urban logistics demand continues to grow, UAV delivery has become a key solution to improve delivery efficiency, reduce traffic congestion, and lower logistics costs. However, to fully leverage the potential of UAV  delivery networks, efficient swarm scheduling and management are crucial. In this paper, we propose a real-time scheduling and management system based on the ``Airport-Unloading Station" model, aiming to bridge the gap between high-level scheduling algorithms and low-level execution systems. This system, acting as middleware, accurately translates the requirements from the scheduling layer into specific execution instructions, ensuring that scheduling algorithms perform effectively in real-world environments. Additionally, we implement three collaborative scheduling schemes involving autonomous ground vehicles (AGVs), unmanned aerial vehicles (UAVs), and ground staff to further optimize overall delivery efficiency. Through extensive experiments, this study demonstrates the rationality and feasibility of the proposed management system, providing a complete solution for urban UAV delivery.

Code: https://github.com/chengji253/UAVDeliverySystem

\end{abstract}

\section{INTRODUCTION}

As urban logistics demand continues to grow, unmanned aerial vehicle (UAV) delivery has emerged as a promising solution to enhance delivery efficiency, alleviate traffic congestion, and reduce logistics costs. An increasing number of companies are deploying UAV delivery networks in urban areas, accelerating technological progress in this field. A crucial factor in achieving high delivery efficiency is the effective management and scheduling of multiple UAVs, ensuring seamless coordination between aerial vehicles and ground personnel.

There has been extensive research on improving the efficiency of UAV delivery systems \cite{9733971}, focusing on the following aspects:
Firstly, optimizing from the perspective of scheduling strategies \cite{SAJID2022109225, murray2015flying, hong2023logistics, das2020synchronized}. Secondly, reducing flight time and avoiding conflicts from a path planning perspective \cite{yan2024optimal, xie2024hybrid, pei2022urban, de2023drones}. 
Additionally, improving energy efficiency is another critical direction \cite{dorling2016vehicle, park2023learning, cho2022multi, huang2022drone}. 
Finally, the efficient use of airspace resources helps increase flight density and reduce conflicts \cite{li2022traffic, pang2024chance, safadi2024integrated}.
Some companies have already made progress in this field, such as Google Wing \cite{wing_website}, Flytrex\cite{flytrex_website}, Amazon \cite{amazon}, Zipline\cite{zipline_website} and Meituan \cite{meituan}.

In the process of commercializing UAV delivery, the most commonly applied is the ``Airport-Unloading Station" model.
In cities, dedicated UAV airports are established as distribution centers. UAVs take off from the airport and fly directly to the unloading station to complete the entire delivery process.
There are many directions to improve the efficiency of the airport model, including flight path planning, UAV swarm scheduling, and the collaboration between ground staff and UAVs (such as takeoff, cargo loading, and battery replacement).
To address this, we need to optimize the entire problem from an operations research perspective, establishing efficient scheduling algorithms to enhance delivery efficiency.

However, applying scheduling algorithms to real-world scenarios still faces numerous challenges.
Firstly, many algorithms often overlook the integration with the underlying execution infrastructure, and the problem modeling includes unrealistic simplifications.
Additionally, many algorithms are designed for offline use, resulting in long calculation times, which are not suitable for real-time scheduling needs.
In practice, we find that the scheduling layer often fails to accurately translate scheduling requirements into specific execution commands.
Furthermore, the scheduling layer is unaware of the exact state of the underlying system and does not know whether the scheduling requirements are correctly executed.
Therefore, there is an urgent need for a real-time management system that bridges the gap between scheduling algorithms and the execution system, providing an effective link for scheduling and execution, ensuring the smooth implementation of scheduling and transport tasks.

To address these challenges, we propose a real-time scheduling and management system. This system serves as middleware for UAV delivery processes, connecting upper-level scheduling requirements with lower-level execution mechanisms, ensuring that scheduling algorithms can run effectively in real-world environments.
Building on this management system, we have also implemented three collaborative scheduling solutions for ground Automated Guided Vehicles (AGVs) and UAVs to ensure efficient coordination between AGVs, UAVs, and ground staff, thereby optimizing overall delivery efficiency.
The main contributions of this study are as follows:
\begin{itemize}
\item We design a  scheduling management system tailored to urban delivery demands, achieving efficient and dynamic UAV delivery scheduling.
\item Based on this management system, we implement three scheduling schemes for AGVs, UAVs, and ground staff, optimizing the coordination between them.
\item The feasibility of the system is verified through experiments, and a practical commercial application solution is provided for drone urban delivery. 
\end{itemize}

\begin{figure}[h]
	\centering
	\includegraphics[scale=0.47]{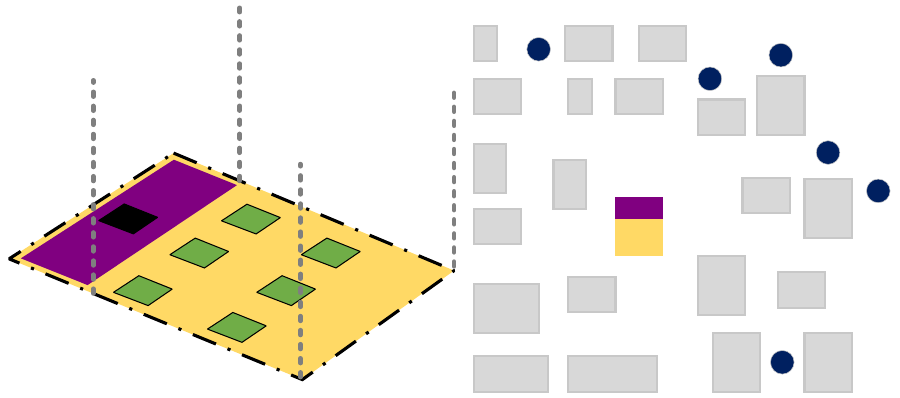}
	\caption{Delivery airport and unloading stations in urban.}
	\label{fig:airport}
\end{figure}

\section{related work}

Improving the efficiency of UAV delivery systems in urban environments can be achieved through various approaches, which can be broadly categorized as follows:

Firstly, optimizing scheduling algorithms plays a significant role in enhancing overall performance. 
Sajid \textit{et al.} \cite{SAJID2022109225} proposed a hybrid optimization framework that combines genetic algorithm and simulated annealing algorithm to solve the path planning and scheduling problems.
Murray \textit{et al.} \cite{murray2015flying} proposed a mathematical model to optimize the cooperative delivery of drones and traditional trucks, and improved the delivery efficiency by optimizing the delivery path and scheduling.
Hong \textit{et al.} \cite{hong2023logistics} proposed a two-stage optimization method to solve the scheduling problem of drones in urban last-mile delivery and improved task allocation and route planning.
Das \textit{et al.} \cite{das2020synchronized} proposed a scheduling mechanism for synchronized drones and delivery trucks to optimize the multi-objective delivery problem.

Second, path planning is another key aspect of optimizing delivery systems.
Yan et al. \cite{yan2024optimal} developed a service model that plans routes through an aerial channel network and employs reactive scheduling to handle failures.
Xie et al. \cite{xie2024hybrid} proposed an AI-based 4D trajectory management system for real-time flight adjustment and airspace conflict resolution.
Pei et al. \cite{pei2022urban} designed an exact algorithm for urban drone delivery considering time windows and spatial conflicts.
Oliveira et al. \cite{de2023drones} investigated UAV collision avoidance in urban environments and evaluated geometric avoidance strategies.

Moreover, energy efficiency is another key factor in UAV delivery optimization.
Dorling et al. \cite{dorling2016vehicle} modeled the joint routing of drones and trucks considering energy and battery consumption.
Park et al. \cite{park2023learning} developed a deep reinforcement learning–based scheduling algorithm to improve multi-drone delivery efficiency while accounting for energy use.
Cho et al. \cite{cho2022multi} proposed a coordinated strategy for hybrid electric trucks and drones to enhance overall energy efficiency.
Huang et al. \cite{huang2022drone} addressed endurance limitations by introducing battery-swap drone stations for last-mile delivery.

Lastly, airspace resource management is vital for scaling UAV delivery systems.
Li et al. \cite{li2022traffic} developed a traffic management framework covering path planning, conflict detection, and resource allocation for low-altitude urban delivery.
Pang et al. \cite{pang2024chance} introduced a chance-constrained optimization model to handle uncertainties such as delays and trajectory deviations.
Safadi et al. \cite{safadi2024integrated} proposed a model-based control strategy to balance demand and supply in low-altitude urban airspace.

\begin{figure}[t]
	\centering
	\includegraphics[scale=0.49]{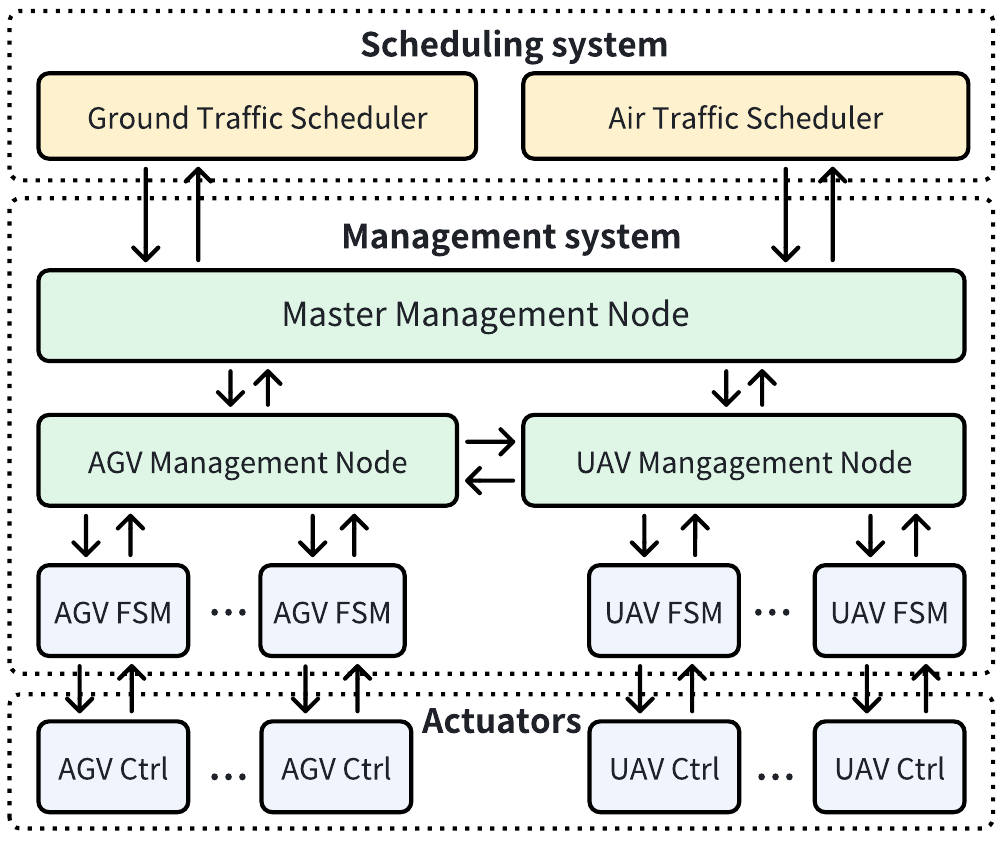}
	\caption{The framework of our system.}
	\label{fig:framework}
\end{figure}

\section{Preliminaries}

Our research is based on the “Airport–Unloading Station” model, where a centralized UAV delivery airport is established in the city. All UAVs take off from the airport, deliver goods to the nearest unloading station, and return for reloading before starting the next delivery task. When a customer places an order, nearby merchants send the goods to the airport, where ground staff handle the loading. Each UAV follows a predefined flight route to an unloading station and then returns. We define the set of unloading stations as $\mathcal{K} = \{1, \dots, K\}$, with each station indexed by $k$. For each $k \in \mathcal{K}$, let $R_{(k)}^{go}$ and $R_{(k)}^{back}$ denote the outbound and return routes, respectively. 
The layout of the airport is shown in Fig. \ref{fig:airport}, where the purple region represents the Ground Work (GW) area for operations like loading, battery replacement, and UAV recovery. The black section within the GW area is the workbench. The orange area denotes the Aviation Work (AW) zone, where UAVs take off and land. To ensure ground staff safety, UAVs do not enter the GW area directly. Instead, AGVs (green area) transfer UAVs between the AW and GW areas. After landing, a UAV is transported by an AGV to the GW area for processing, then returned to the AW area for its next flight. The set of UAVs is denoted by $\mathcal{N} = \{1, \dots, N\}$ (indexed by $n$), and the set of AGVs is $\mathcal{M} = \{1, \dots, M\}$ (indexed by $m$). The black circles in Fig. \ref{fig:airport} represent the unloading stations in the city, where goods are unloaded and made available for customer pickup.

\begin{figure*}[t]
	\centering
	\includegraphics[scale=0.35]{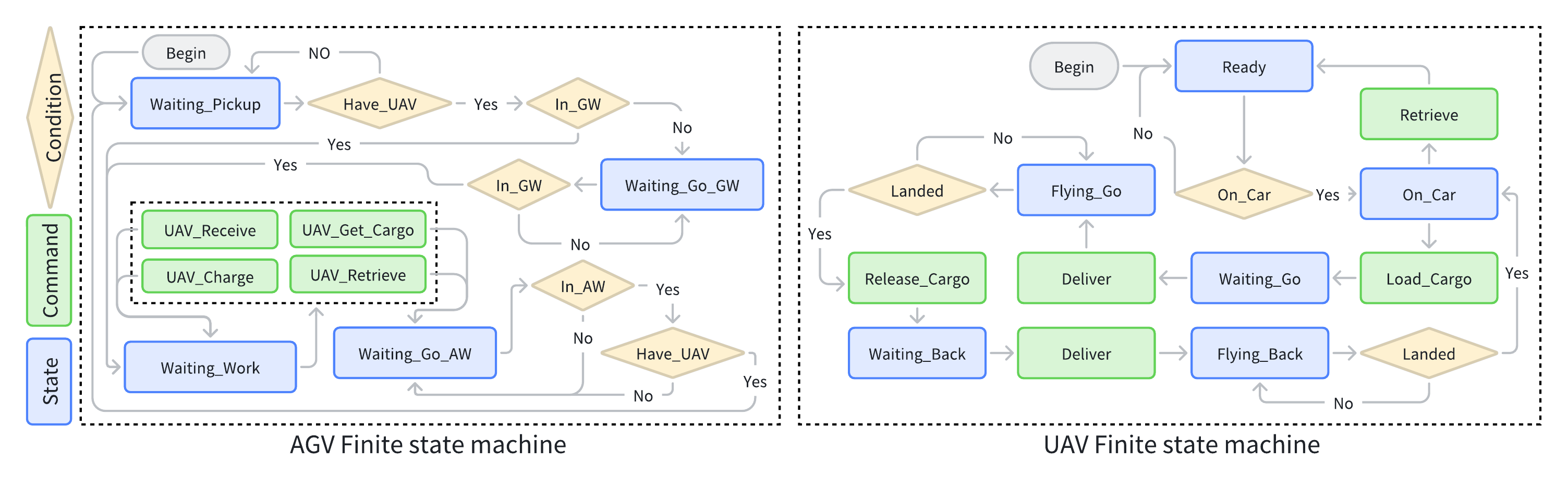}
	\caption{Illustration of UAV FSM and AGV FSM. The blue part indicates the state. The green part indicates the command. The yellow part indicates the condition.}
	\label{fig:uav fsm}
\end{figure*}
\section{Management system}

Our management system functions as middleware between the execution infrastructure and the high-level scheduling layer, as shown in Fig.~\ref{fig:framework}. 
It comprises three core components—Master Management Node, AGV Management Node, and UAV Management Node—each operating as an independent real-time process and collaborating through information exchange.

\subsection{Management Nodes}

AGV Management Node:
All AGVs are managed through a combination of multithreading and finite state machines (FSMs). Each AGV is controlled by an independent FSM thread that updates its state at a fixed frequency, ensuring precise and highly responsive control. The node receives commands from the master node, forwards them to the corresponding FSM, and continuously publishes standardized status information.
UAV Management Node:
Similarly, the UAV management node adopts multithreading and FSMs, with each UAV running independently to guarantee real-time and autonomous task execution. The node receives commands from the master node, relays them to the respective FSM, and continuously publishes UAV status updates.
Master Management Node:
As the core of the system, the master node connects the scheduling layer with the execution layer. It subscribes to AGV and UAV status updates, provides the aggregated information to the scheduler for decision-making, and distributes scheduling results to the corresponding management nodes for execution.
The system adopts a distributed architecture in which each vehicle is managed by an independent thread, offering:
(1) Fault Isolation: A failure in one vehicle does not affect others.
(2) Real-Time Responsiveness: Continuous monitoring and dynamic task adjustment.
(3) Scalability: New vehicles can be seamlessly added by assigning new threads.

\subsection{UAV Finite state machine}

Fig.~\ref{fig:uav fsm} illustrates the state machine of the UAV. In the diagram, blue boxes represent states, green boxes indicate commands, and yellow boxes denote conditional checks. The UAV operates through six states: \textsf{Ready}, where it remains idle at the airport awaiting loading; \textsf{On\_Car}, indicating it has been mounted on an AGV; \textsf{Waiting\_Go}, where it awaits the takeoff command after loading cargo; \textsf{Flying\_Go}, when it is en route to the unloading station; \textsf{Waiting\_Back}, where it waits for the return command after unloading; and \textsf{Flying\_Back}, during which it flies back to the airport.
The UAV responds to three main commands: \textsf{Delivery}, which triggers takeoff to the delivery point; \textsf{Release\_Cargo}, which initiates cargo unloading (only allowed when landed); and \textsf{Load\_Cargo}, which loads goods onto the UAV.
State transitions rely on four condition checks: \textsf{Landed} verifies if the UAV has landed, \textsf{On\_Car} checks whether it is mounted on an AGV, \textsf{Get\_Cargo} confirms if cargo has been loaded, and \textsf{Retrieved} determines whether the UAV has been recovered.

\subsection{AGV Finite state machine}

The AGV is responsible for transporting UAVs between the AW and GW areas, managing UAV takeoff and landing, and assisting with loading and unloading operations. To support these tasks, the AGV is designed with four working states and corresponding task commands.
The four AGV working states are as follows: \textsf{Waiting\_Go\_GW} indicates the AGV is preparing to head to the GW area after receiving a landed UAV in the AW area; \textsf{Waiting\_Pickup} refers to the state when the AGV is waiting in the AW area for a UAV to land, without a UAV onboard; \textsf{Waiting\_Working} occurs when the AGV, carrying a UAV, arrives at the GW area and awaits ground staff to perform operations such as cargo loading or battery replacement; \textsf{Waiting\_Go\_AW} follows task completion and indicates the AGV is ready to return to the AW area with the UAV.
The AGV executes four main commands: \textsf{UAV\_Get\_Cargo} instructs the UAV to load cargo at the GW workbench; \textsf{UAV\_Receive} directs the AGV to load the UAV upon arrival at the GW; \textsf{UAV\_Retrieve} removes the UAV from the AGV, returning it to idle; and \textsf{UAV\_Charge} initiates a battery replacement for the UAV.
AGV state transitions depend on three condition checks: \textsf{Have\_UAV} verifies if the AGV currently carries a UAV; \textsf{In\_GW} checks whether it is located in the GW area; and \textsf{In\_AW} determines if it is in the AW area.

\subsection{State Transition Process}

We now describe a full delivery cycle using the UAV and AGV finite state machines. Initially, the UAV is in the \textsf{Ready} state and the AGV in \textsf{Waiting\_Pickup}. The AGV moves to the GW area and, upon receiving the \textsf{UAV\_Receive} command, loads the UAV, transitioning the UAV to \textsf{On\_Car} and the AGV to \textsf{Waiting\_Working}.
Next, both receive the cargo loading command. Once loading is complete, the UAV enters \textsf{Waiting\_Go}, and the AGV changes to \textsf{Waiting\_Go\_AW}. The AGV transports the UAV to the AW area, where it issues the \textsf{Delivery} command. The UAV takes off, flies to the unloading station, and upon arrival, receives the unload command and enters \textsf{Waiting\_Back}.
The AGV then sends another \textsf{Delivery} command for the UAV to return. After landing back at the airport, the UAV re-enters \textsf{On\_Car}, and the AGV transitions to \textsf{Waiting\_Go\_GW}. It carries the UAV to the GW area for reloading, completing one delivery cycle.

\section{Scheduling system}

In commercial aviation, takeoff involves several coordinated steps—clearances for departure, taxiing, and takeoff, followed by post-takeoff coordination—managed by air traffic control.
Inspired by this process, our system autonomously manages UAV delivery without human intervention. It employs two coordinated schedulers for ground and air traffic, supported by the previously introduced management framework, enabling flexible and efficient scheduling strategies.
The following section presents the design of these schedulers.

\begin{figure}[t]
	\centering
	\includegraphics[scale=0.75]{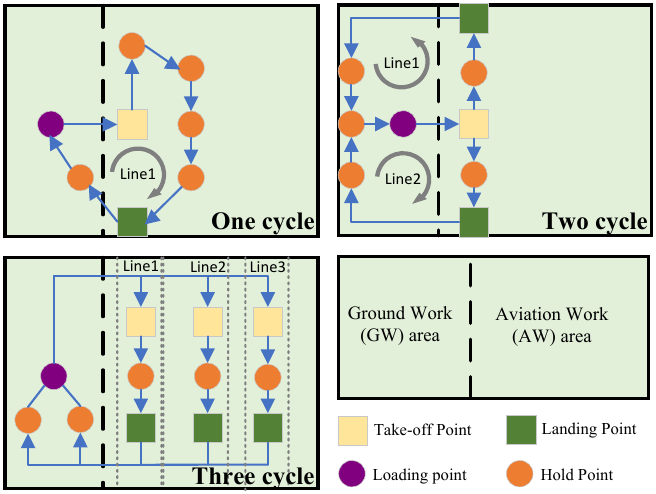}
	\caption{AGV scheduling process of three schemes}
	\label{fig:jichang}
\end{figure}

\subsection{Ground traffic scheduler}

The ground traffic scheduler manages AGVs transporting UAVs between the AW and GW areas. As shown in Fig.~\ref{fig:jichang}, it supports three scheduling schemes: One-Cycle, Two-Cycle, and Three-Cycle, each defining key nodes such as the Takeoff Point (yellow), Hold Point (blue), Loading Point (purple), and Landing Point (green).
One-Cycle: All six AGVs operate in a single loop with one takeoff and one landing point. After loading, an AGV moves to the takeoff point, then cycles through hold points before reaching the landing point for UAV recovery.
Two-Cycle: Two loops share one takeoff point and have two separate landing points, with three AGVs assigned to each loop.
Three-Cycle: Three independent loops each have their own takeoff and landing points, with two AGVs per loop.
All schemes follow an assembly line-like process, ensuring that while one AGV heads to the GW area, another is ready at the landing point. This cyclic design minimizes UAV wait times and ensures efficient handover between air and ground operations.

\begin{figure*}[t]
	\centering
	\includegraphics[scale=0.88]{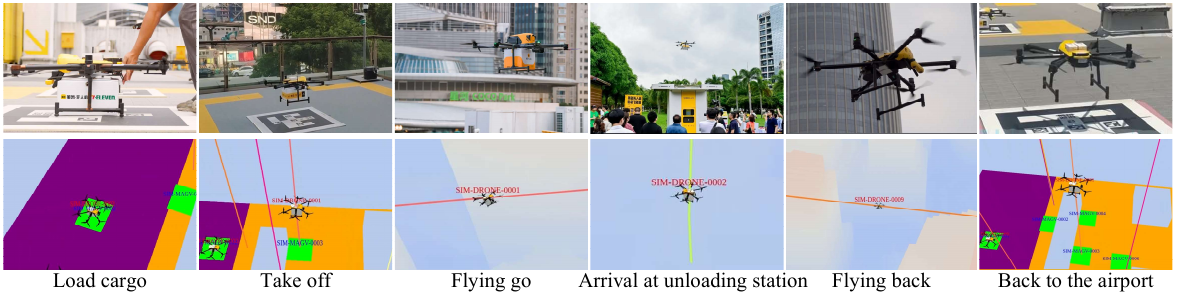}
	\caption{UAV delivery workflow: real-world and simulation visualization.}
	\label{fig: UAV-Delivery}
\end{figure*}

\subsection{Air traffic scheduler}

The air traffic scheduler determines whether a UAV can take off or return and issues corresponding flight route instructions.
When an AGV transports a UAV to the takeoff point, the UAV sends a takeoff request. The scheduler then checks for approval and, if granted, assigns the outbound route $R_{(k)}^{go}$. Similarly, upon reaching the unloading station, the UAV sends a return request. The scheduler then assigns the return route $R_{(k)}^{back}$ and a landing point.
To approve a takeoff, the scheduler considers potential conflicts at the unloading station. Let $T_{(n)}^{go\_req}$ be the time the $n$-th UAV requests takeoff, and $t_{(n)}^{go}$ the estimated flight time to station $k$. Then the UAV's arrival time is:
$T_{(n)}^{arri} = T_{(n)}^{go\_req} + t_{(n)}^{go} $.
Let $\mathcal{N}_k^{go}$ be the set of UAVs en route to station $k$. The takeoff is allowed only if:
\begin{equation}
|T_{(n_i)}^{arri} - T_{(n_j)}^{arri}| > t^{go\_gap}, \quad \forall n_j \in \mathcal{N}_k^{go}
\end{equation}
ensuring a minimum time gap $t^{go\_gap}$ between UAV arrivals.
For returns, let $T_{(n)}^{back\_req}$ be the request time and $t_{(n)}^{back}$ the flight time. Then:$
T_{(n)}^{land} = T_{(n)}^{back\_req} + t_{(n)}^{back}$.
A return is approved only if an AGV can arrive at the landing point before the UAV:
\begin{equation}
T_{(m)}^{land} < T_{(n)}^{land}
\end{equation}
The scheduler estimates $T_{(m)}^{land}$ based on the AGV’s current state. If multiple AGVs or lines satisfy the condition, the one with the fewest reservations is chosen.

By dynamically evaluating UAV requests and AGV availability, the scheduler ensures safe and efficient coordination of takeoff and landing operations.

\noindent \textbf{Remark}:
In the air traffic scheduler section, we assume the flight routes $ R_{(k)}^{go} $ and \( R_{(k)}^{back} \) are predefined.
In real urban delivery scenarios, planning safe and efficient UAV routes between airports and unloading stations is a complex problem involving airspace utilization and collision avoidance.
As this is beyond the scope of our work, we simplify the study by assuming that feasible flight paths are pre-planned and available.

\section{Experiments}

This section presents the experimental analysis of our system. Our experiments were conducted on a PC with an Intel Core i7-12700 CPU and 32GB of RAM. 
The simulation platform comprises four functional Docker modules, each handling a specific task. The AGV Docker simulates AGV low-level control, where each AGV runs an independent real-time controller. The UAV Docker manages UAV control using a PX4-like flight controller with MPC-based trajectory tracking. The Scenario Docker oversees the simulation environment, performing collision detection and coordinating interactions among UAVs, airports, and unloading stations. The Scheduling Docker executes the proposed management system, responsible for global task allocation and coordination. All modules communicate through a Docker user-defined network using ROS for task publishing, state transmission, and system coordination. This modular design, with UAV and AGV controllers separated into distinct containers, provides better scalability and more realistic emulation of real-world operations.

Fig.~\ref{fig: UAV-Delivery} illustrates the complete UAV delivery workflow—from AGV transport and takeoff to delivery and return.
Experimental parameters are as follows:
Cargo loading: 10s; battery replacement: 10s; cargo unloading: 3s.
Minimum distance: UAVs 5m, AGVs 3m; max speeds: UAVs 10 m/s, AGVs 1.5 m/s.
Node frequencies: master 2 Hz AGV /UAV management 10 Hz; FSMs 10 Hz each.

In terms of goods orders, we generated a delivery order list in advance.
Each order has four time attributes:
(1) \textit{OrderT}: The time when the customer places the order.
(2) \textit{FinishT}: The time when the order is successfully delivered.
(3) \textit{BetterT}: The ideal delivery time, typically based on customer expectations or system optimization.
(4) \textit{TimeOut}: The maximum allowable delivery time; if exceeded, the order is considered overdue.
We will give a score based on the time when the UAV delivers the goods.
The scoring rules for the system are governed by the following piecewise function:
\begin{equation}
\begin{cases}
	100, & \text{if } \textit{OrderT} \leq \textit{FinishT} \leq \textit{BetterT}, \\
	100 \times \frac{\textit{TimeOut} - \textit{FinishT}}{\textit{TimeOut} - \textit{BetterT}}, & \text{if } \textit{BetterT} < \textit{FinishT} \leq \textit{TimeOut}, \\
	-100 \times \frac{\textit{FinishT} - \textit{Timeout}}{\textit{TimeOut} - \textit{BetterT}}, & \text{if } \textit{FinishT} > \textit{TimeOut}.
\end{cases}
\label{score func}
\end{equation}
If the \textit{FinishT} falls between \textit{OrderT} and \textit{BetterT}, the score is a fixed value of 100.  
If the \textit{FinishT} falls between \textit{BetterT} and \textit{TimeOut}, the score is calculated based on the Equ.(\ref{score func}), where earlier deliveries earn higher scores.  
If the \textit{FinishT} exceeds \textit{TimeOut}, the score is negative, and the later the delivery, the higher the penalty.  

In the experiment, we tested the impact of varying the number of UAVs (4, 6, 8, 10, 12, 14, 16) on the system's performance. For each configuration, the system was run for one hour, recording the number of deliveries and scores to evaluate system performance.
Fig. \ref{fig: UAV-Delivery} shows the scene diagram of the UAV delivery. 
We can see a demonstration of the cargo loading, taking off, departing, arriving at the unloading station, returning and finally landing at the airport.
For a more detailed UAV delivery demonstration, please refer to our supplementary video.

\begin{figure}[t]
	\centering
	\begin{minipage}{0.23\textwidth}
		\centering
		\includegraphics[width=\linewidth]{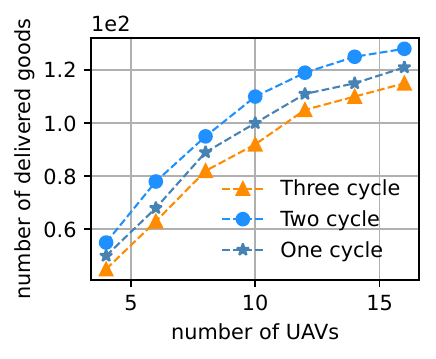}
		\label{fig:numbergoods-uavs}
	\end{minipage}
	\begin{minipage}{0.23\textwidth}
		\centering
		\includegraphics[width=\linewidth]{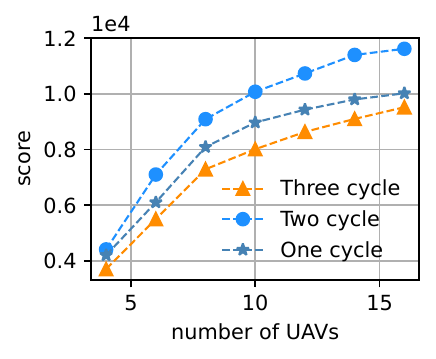}
		\label{fig:score-uavs}
	\end{minipage}
	\caption{The number of delivered goods and their corresponding scores  w.r.t. to the number of UAVs.}
	\label{fig:score-goods}
\end{figure}

\begin{figure}[t]
	\centering
	\begin{minipage}{0.23\textwidth}
		\centering
		\includegraphics[width=\linewidth]{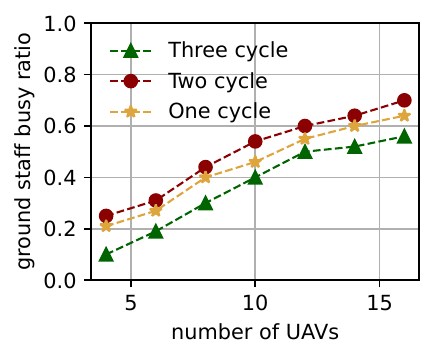}
		\label{fig:ratio-work}
	\end{minipage}
	\begin{minipage}{0.23\textwidth}
		\centering
		\includegraphics[width=\linewidth]{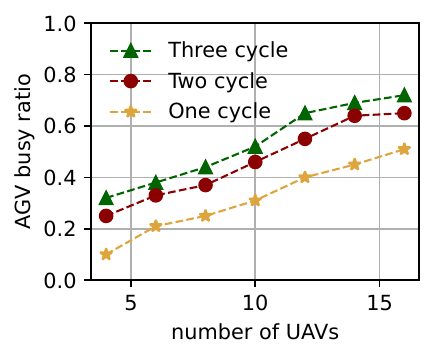}
		\label{fig:ratio-agv}
	\end{minipage}
	\caption{Ground staff and AGV busy ratios w.r.t. to the number of UAVs.}
	\label{fig:ratio}
\end{figure}

Fig. \ref{fig:score-goods} shows the number of goods delivered and the scores obtained after a one-hour experiment using three different scheduling schemes with different numbers of drones.
It can be observed that: 
(1) As the number of UAVs increases, the number of deliveries and scores of the three scheduling schemes will increase.
The Two-Cycle received the highest score, followed by the One cycle, and the Three-Cycle least.
(2) Each time the number of UAVs increases, the percentage of increase in the score will gradually decrease. This shows that the increase in UAVs will improve the efficiency of transportation, but the efficiency improvement has an upper limit.
(3) The average delivery score is around 80 points, which means that many orders cannot be delivered before \textit{BetterT}.

We define the AGV as being in a “free” state when it is idle and waiting for instructions, while all other states are considered “busy.” The time proportions of AGVs and ground staff in different states within one hour were recorded accordingly.
Fig. \ref{fig:ratio} illustrates the busy ratios of AGVs and ground staff under three scheduling schemes as the number of UAVs increases. The results show that:
(1) Increasing the number of UAVs raises both AGV and ground staff busy ratios;
(2) The Two-cycle scheme yields the highest ground staff busy ratio, while the AGV busy ratio is moderate;
(3) The Three-cycle scheme results in the lowest ground staff busy ratio but the highest AGV busy ratio.
This is because in the Three-cycle scheme each AGV spends more time traveling along its route, while the Two-cycle scheme achieves higher ground utilization with a lower AGV workload.

Combining Fig. \ref{fig:score-goods} and Fig. \ref{fig:ratio}, we can draw the following conclusions:
(1) The ranking of ground staff busy ratios aligns with the delivery scores, indicating that higher utilization means more cargo loaded and thus higher scores.
(2) The Two-cycle scheme achieves the highest score with relatively low AGV utilization, making it the most efficient.
(3) In the Three-cycle scheme, AGVs spend excessive time traveling, reducing overall efficiency.
(4) The One-cycle scheme shows low AGV utilization due to long waiting times; although the Three-cycle scheme reduces waiting, its longer travel distance still leaves room for improvement.

An efficient scheduling scheme should balance AGV and ground resource utilization through well-designed takeoff/landing points, waiting areas, and route planning. Precise coordination between UAV landings and ground operations is also crucial for smooth performance. Although we tested only three schemes, better designs are certainly possible.

\section{Conclusion}

This paper presents a real-time scheduling and management system for urban UAV delivery, effectively bridging the gap between high-level scheduling algorithms and low-level execution systems. Through three collaborative scheduling schemes, the system optimizes the coordination of UAVs,  AGVs, and ground staff, improving delivery efficiency. Experimental results validate the system's rationality and feasibility in urban logistics scenarios.

\section{Acknowledgments}
We would like to thank Meituan Academy of Robotics Shenzhen for supporting this research.

\balance

\bibliography{citepaper_all}

\end{document}